\documentclass[10pt,twocolumn,letterpaper]{article}

\usepackage{wacv}              

\usepackage{graphicx}
\usepackage{amsmath}
\usepackage{amssymb}
\usepackage{booktabs}
\usepackage{algorithm}
\usepackage{algorithmic}

\usepackage[pagebackref,breaklinks,colorlinks]{hyperref}

\usepackage[capitalize]{cleveref}
\crefname{section}{Sec.}{Secs.}
\Crefname{section}{Section}{Sections}
\Crefname{table}{Table}{Tables}
\crefname{table}{Tab.}{Tabs.}

\def\wacvPaperID{1559} 
\def\confName{WACV}
\def\confYear{2024}

\begin{document}

\title{HDformer: A Higher-Dimensional Transformer for Detecting Diabetes Utilizing Long-Range Vascular Signals}

\author{Ella Lan\\
}
\maketitle

\begin{abstract}
   Diabetes mellitus is a global concern, and early detection can prevent serious complications. 50\% of people with diabetes live undiagnosed, disproportionately afflicting low-income groups. Non-invasive methods have emerged for timely detection; however, their limited accuracy constrains clinical usage. In this research, we present a novel Higher-Dimensional Transformer (HDformer), the first Transformer-based architecture which utilizes long-range photoplethysmography (PPG) to detect diabetes. The long-range PPG maximizes the signal contextual information when compared to the less-than 30 second signals commonly used in existing research. To increase the computational efficiency of HDformer’s long-range processing, a new attention module, Time Square Attention (TSA), is invented to reduce the volume of tokens by more than 10x, while retaining the local/global dependencies. TSA converts the 1D inputs into 2D representations, grouping the adjacent points into a single 2D token. It then generates dynamic patches and feeds them into a gated mixture-of-experts (MoE) network, optimizing the learning on different attention areas. HDformer achieves state-of-the-art results (sensitivity 98.4, accuracy 97.3, specificity 92.8, AUC 0.929) on the standard MIMIC-III dataset, surpassing existing research. Furthermore, we develop an end-to-end solution where a low-cost wearable is prototyped to connect with the HDformer in the Cloud via a mobile app. This scalable, convenient, and affordable approach provides instantaneous detection and continuous monitoring for individuals. It aids doctors in easily screening for diabetes and safeguards underprivileged communities. This minimizes treatment delays and saves lives. The enhanced versatility of HDformer allows for efficient processing and learning of long-range signals in general one-dimensional time-series sequences, particularly for all biomedical waveforms.  
\end{abstract}

\section{Introduction}

Diabetes mellitus is a clinical condition that results in a high amount of glucose in the blood due to a lack of insulin in the body— otherwise known as insulin resistance \cite{martin1992role}. Diabetes increases the risk of complications in nearly every organ system, leading to conditions such as coronary heart disease, kidney failure, blindness, and stroke. According to the World Health Organization, about 537 million people worldwide are diagnosed with diabetes, and these statistics have a notably disproportionate impact on lower-income communities.

Diabetes is a “silent killer”; it is often overlooked until its progression into critical stages. The lack of obvious symptoms at onset prevents diabetic patients from being treated until later stages when the patients' blood sugar is uncontrollable and acutely above the standard. According to data from the International Diabetes Federation, almost 50\% of people with diabetes are unaware of their diagnosis and its risks to their health \cite{ogurtsova2022idf}, hence leaving the disease untreated. 

Thus, early detection is critical to prevent long-term complications and reduce mortality rates. 


To diagnose diabetes, clinics use invasive or semi-invasive methods. However, such treatments are expensive, time-consuming, and inconvenient for patients. To overcome these limitations, research on non-invasive methods has emerged. A continuous, noninvasive, painless, easy, and low-cost solution can improve patient adherence to routine blood glucose monitoring, which may result in earlier diabetes detection. 

\begin{figure}[t]
  \centering
  \includegraphics[width=0.9\linewidth]{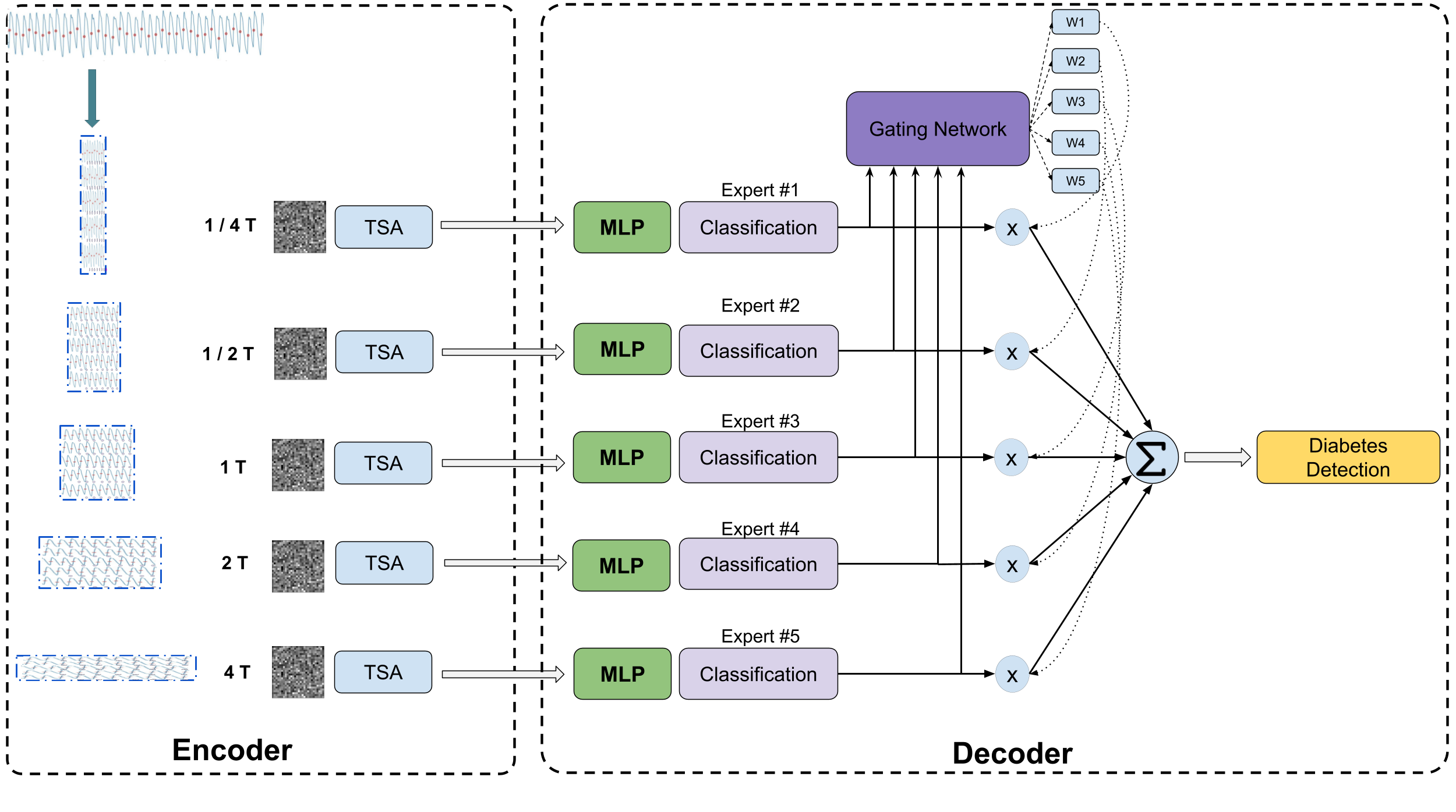}
   \caption{HDformer Architecture.}
   \label{fig:figure2}
\end{figure}

Photoplethysmography (PPG) is an optically obtained signal that can be used to detect blood volume changes in the microvascular bed of tissues. PPG can extract various pieces of cardiovascular-related information \cite{elgendi2019use}. Clinics have noted that diabetes is linked to vascular changes. In particular, diabetic groups often exhibit signs of reduced heart rate variability and an elevated resting heart rate. Many key markers for diabetes are reflected in the PPG waveform. For this reason, PPG is often considered when measuring blood glucose estimation and diabetes detection \cite{thayer2006beyond}.

However, at the same time PPG is often sensitive and noisy: it is easily affected by motion, light, skin type, etc., and this limits its  applicabilities in the real-world. The usability of PPG in clinics can be further enhanced by deep learning, via an automated, intricate analysis of the contextual relationships both between and within the waveforms. The development of such technologies capable of detecting the onset of diabetes can lead to large-scale prevention. However, the accuracy and general applicability of these non-invasive approaches have not proven to be competitive with current invasive methods. 

\begin{figure}[t]
  \centering
  \includegraphics[width=0.9\linewidth]{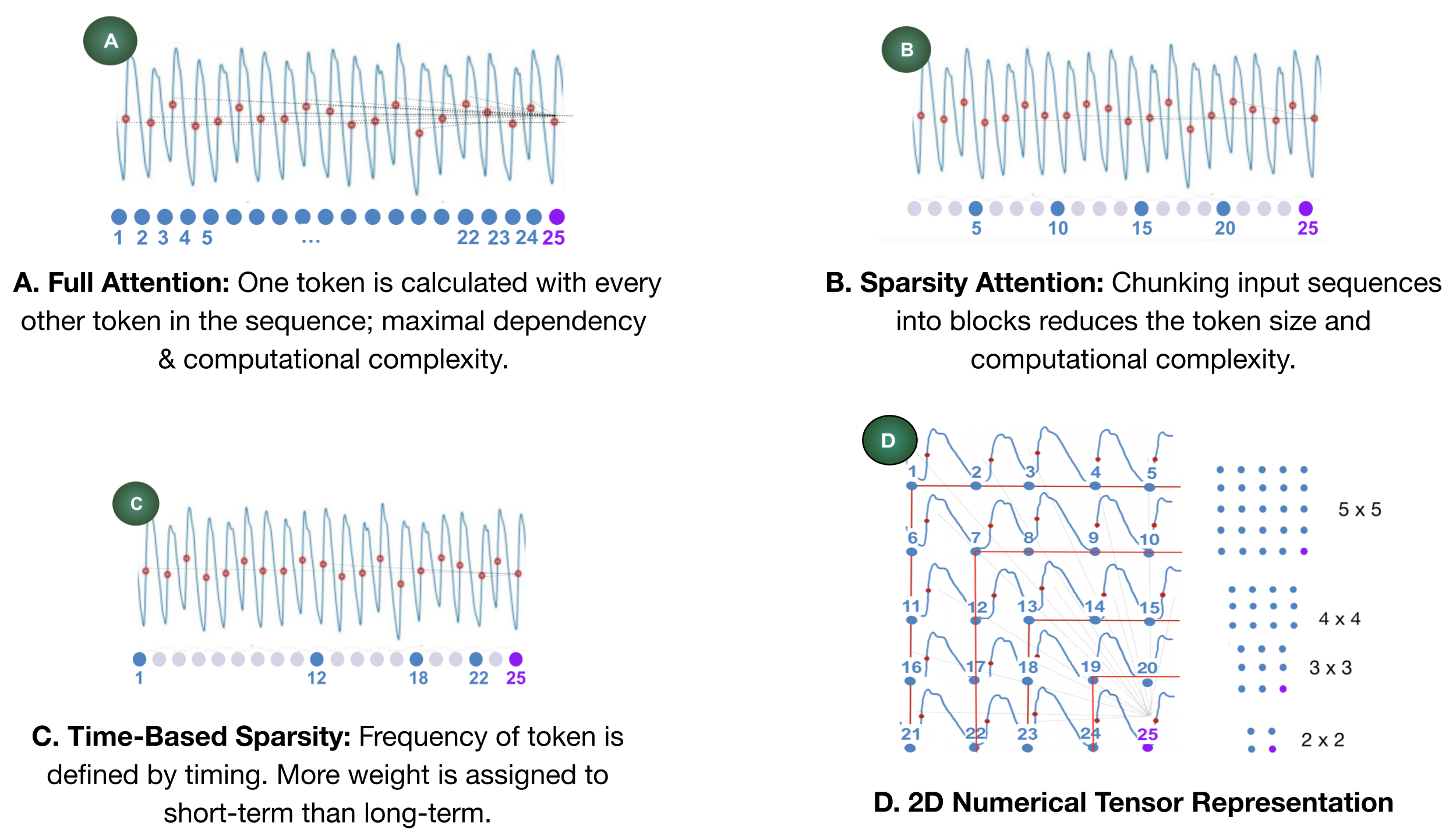}
   \caption{Various Attentions Comparison and TSA.}
   \label{fig:figure3}
\end{figure}

To address the accuracy gap, the present research proposes long-range PPG waveforms (10+ min) as the input, unlike the existing research which analyzes signals of less than 30 seconds. Long-range vascular signals have richer features that enable precisely classifying diabetes. In this study, we propose a Higher-Dimensional Transformer (HDformer), capturing the global representation and long-distance feature dependencies among PPG waveforms via attention modules. A new Time-Square Attention (TSA) is created to aggregate one-dimensional (1D) dependencies from two-dimensional (2D) representations. The proposed ML model has achieved SOTA results on the standard MIMIC-III dataset. The contributions of this paper include:

\begin{itemize}
    \item A novel, scalable, non-invasive, end-to-end solution for using long-range vascular signals (PPG) to detect diabetes, achieving SOTA results. As the hardware component, we also created an AI-based PPG wearable.

    \item A Transformer-based deep-learning architecture HDformer, to perform long-range biomedical waveforms processing.

    \item A proposed attention module TSA to capture 1D dependencies from 2D representations, adaptable as inputs for existing 2D Transformer models, while applying a gated network of mixture-of-experts for the dynamic patch size of each 2D shape.

    \item A deep-learning-based, in-depth, long-range data analysis on the blood volume changes (measured from PPG) for diabetes detection, and an introduction of the multi-modal extension by reconstructing ECG from PPG, reducing the required data input length from 10 min to 5 min.

    \item A general Transformer-based framework capable of time-series learning and prediction for 1D long-range sequences, especially for all biomedical waveforms.
\end{itemize}

\begin{figure*}[t]
  \centering
  \includegraphics[width=0.9\linewidth]{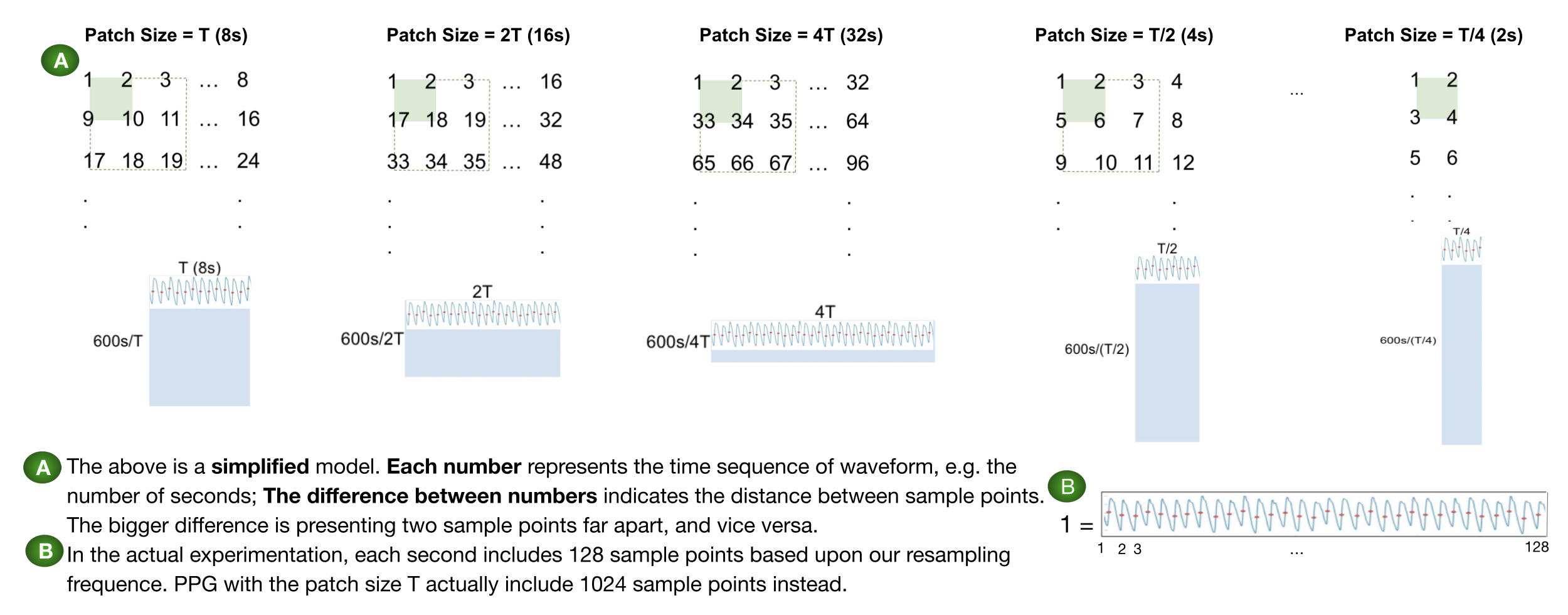}
   \caption{Concept of Dynamic Patch Sizes in TSA.}
   \label{fig:figure4}
\end{figure*}

\section{Related Work}

\subsection{Photoplethysmography (PPG)}

PPG is a commonly used digital biomarker in cardiovascular disease (CVD) analysis. Since PPG is measured by non-invasive methods, it has recently been introduced in tasks like blood glucose estimation and diabetes detection through machine-learning approaches. 

One of the first studies in this area was by ~\cite{moreno2016type}, who used the inverse Fourier transform to extract features to feed into several machine-learning models. Further, ~\cite{hettiarachchi2019machine} identified features related to diabetes from PPG and established the feasibility of prediction with its linear discriminant analysis (LDA), and ~\cite{qawqzeh2020classification} developed logistic regression modeling to use PPG to classify diabetes. However, to obtain reliable results, these methods required an abundant amount of attention to data processing for feature extraction. Additionally, each study collected its own datasets, producing a lack of result standardization among the results. 

These limits make the traditional machine-learning methods challenging to scale to broader usage. The recent rise of deep learning has led to the application of convolutional neural networks (CNN) in predicting diabetes using PPG. ~\cite{avram2020digital} used smartphone-based PPG signals and CNNs. ~\cite{panwar2020cardionet} presented a reconfigurable deep-learning framework, combining CNNs and the inherent capabilities of PPG feature extraction. ~\cite{wang2020igrnet} and ~\cite{srinivasan2021deep} proposed combining 3D CNN models— one taking ECGs and another taking PPGs— with other information like age, gender, and the presence of hypertension. However, their training required larger datasets to be generated by themselves. Additionally, because CNNs are locality-sensitive, the accuracy of these models is limited to 70\%-80\%, lower than that of feature-extraction-based machine-learning models. In our research, we chose PPG over ECG because it enables continuous monitoring, and Transformers over CNNs to capture global contextual information and long-range dependencies for classifying diabetes.

\subsection{Long-Range Transformers}

Transformers have become a fundamental building block for a variety of state-of-the-art natural-language-processing tasks, including training of large language models (LLMs) like ChatGPT and PALM-2. Although Transformers ~\cite{vaswani2017attention} originated in the world of natural language processing (NLP), it has also become prevalent in the field of computer vision, surpassing many CNN-based models in performing tasks such as image classification and segmentation ~\cite{dosovitskiy2020image} ~\cite{carion2020end}. Much of the sucess of Transformers comes from their self-attention mechanism, which not only simplifies the architectural complexity by removing convolutions and allows models to interpret the intricate relationships among tokens, and capture global contextual information for both short-range and long-range relationships. 

Recent studies suggest that enhancing architectures, such as Transformers, can make them better suited for analyzing long-range data. Such improvements are achieved by optimizing components like self-attention mechanisms and improving the efficiency of memory usage. These proposals, built on top of the vanilla Transformer, include the memory optimization-based LongFormer ~\cite{beltagy2020longformer}, lower-dimensional-representation-based LinFormer ~\cite{wang2020linformer}, recurrence-based Transformer XL ~\cite{dai2019transformer}, downsampling-based Informer ~\cite{zhou2021informer}, and learnable-patterns-based Reformer ~\cite{kitaev2020reformer}, etc. However, these models still are limited in processing data, e.g. the token limits are 4k for GPT-3.5 and 8k+ for GPT-4 \cite{chen2023lm4hpc}.

In the present study, we propose a new Transformer architecture, HDformer, which processes the 1D PPG waveforms into 2D representations via a novel attention model TSA and an efficient tokenization mechanism, optimizing model efficiency while retaining the key information in the signals.

\section{Methods}

\subsection{Long-Range Vascular Signals}

Due to the complexity of CNNs, PPG signals with a duration of up to 30s are commonly used; however, this restricted duration limits model performance. Long-range PPG includes richer features and adds long-term changes and relationships for models to analyze and helps separate stable waveform variation from one-time noise. Additionally, metrics like heart rate variability, which is typically measured by PPG and is linked to glucose levels should be analyzed for at least 5 min \cite{shaffer2017overview}. Long-range PPG helps provide a complete picture of heart rate variability, and its long-range data collection can contain more long-distance features that are missed in the short-distance PPG. To capture long-distance features, we propose a new Transformer, Higher-Dimensional Transformer (HDformer) to process the long-range PPG waveforms for classifying diabetes. Our Transformer-based method can model more complex relationships and capture richer contextual information by taking 10+ min of PPG signals as the input.

\subsection{Design}

\begin{algorithm}[tb]
\caption{Time Square Attention Algorithm}
\label{alg:algorithm1}
\textbf{Input}: 1D PPG Waveforms\\
\textbf{Parameter}: Original Patch Size T, Targeted 2D Representation Size (N, T), Dynamical Patch Size D, Neural Network Layer Config\\
\textbf{Output}: Diabetes Classification
\begin{algorithmic}[1] 
\STATE Initial patch partition.
\WHILE{current 2D representation size is smaller than (N, T)}
\IF {$\mbox{D} > $\mbox{T}}
\STATE Fetch the length of D from PPG waveforms 
\STATE Perform down-sampling from D to T
\STATE Add into 2D representation
\ENDIF
\IF {$\mbox{D} <= $\mbox{T}}
\STATE Fetch the length of D * N from PPG waveforms
\STATE Partition into N patch of the D size
\STATE Add into 2D representation
\ENDIF
\ENDWHILE
\STATE Feed 2D representation into Transformer Encoder (in our study, we take Swin Transformer)
\STATE Add MLP layer for diabetes classification
\STATE \textbf{return} Diabetes prediction
\end{algorithmic}
\end{algorithm}

HDformer uses an encoder/decoder-based architecture. In the encoder phase, raw PPG signals are de-noised and normalized in a pre-processing module. After standard segmentation, each sequence represents a 10 min PPG waveform. A hierarchical design is then structured as follows: a patch partition operation is taken to create patches of the PPG waveforms, which are then constructed into 2D waveform representations with different shapes. This design is expandable to include more layers for various patch shapes. Each encoder contains a TSA which processes the 2D representations and can easily be inserted into existing 2D Transformers (e.g., ViT ~\cite{dosovitskiy2020image} or Swin ~\cite{liu2021swin}). In the decoder phase, an MLP-based classification is performed on each model. The predictions from these models (experts) are then fed as decoders into a gated network following the mixture-of-experts framework. Finally, the model corroborates all outcomes before outputting a diabetes detection value, as illustrated in Figure~\ref{fig:figure2}.


\begin{table}
  \begin{center}
  \begin{tabular}{@{} c c c c c @{}}
    \toprule
    Approaches & Sen. & Acc. & Spe. & AUC \\
    \midrule
    Moreno \cite{moreno2016type} & 80.0 & 70.0 & 48.0 & - \\
    Reddy \cite{reddy2017perdmcs} & 84.0 & 82.0 & 80.0 & - \\
    Hettiarachchi \cite{hettiarachchi2019machine} & - & 83.0 & - & - \\
    Qawqzeh \cite{qawqzeh2020classification} & 70.0 & 92.3 & 96.0 & - \\
    Avram \cite{avram2020digital} & 75.0 & 76.7 & 65.5 & 0.770\\
    Wang ~\cite{wang2020igrnet} & 80.8 & 77.8 & 77.5 & 0.770\\
    Srinivasan \cite{srinivasan2021deep} & 76.7 & 76.3 & 76.1 & 0.830\\
    HDformer & \textbf{98.4} & \textbf{97.3} & \textbf{92.8} & \textbf{0.929}\\
    \bottomrule
  \end{tabular}
  \end{center}
  \caption{Comparison of Our Results with Related Work.}
  \label{tab:sota}
\end{table}

\subsection{Time Square Attention (TSA)}

The attention calculation is described as 

\[ Attention(Q, K, V) = Softmax \left( \frac{Q K^{T}}{\sqrt{d_{k}}} \right) V \]

where Q, K, V denote Query, Key, and Value, as defined by Transformer’s self-attention.

While much of the success of Transformers relies on their self-attention module, its computational complexity and memory usage grow quadratically along with the length of the sequence. Thus, it is inefficient and infeasible for standard Transformers to process the long-range data. Hence, TSA handles the PPG waveforms as 2D representations rather than as 1D sequence. We create a 2D representation by first partitioning the 1D waveform into patches before then constructing these patches into 2D data, inspired by the time-series nature of the PPG waveforms’ recurring patterns. To address the limitatiosn of self-attention on long-range data, various attention models are illustrated and compared in Figure~\ref{fig:figure3}.

\begin{figure}[t]
  \centering
  \includegraphics[width=0.9\linewidth]{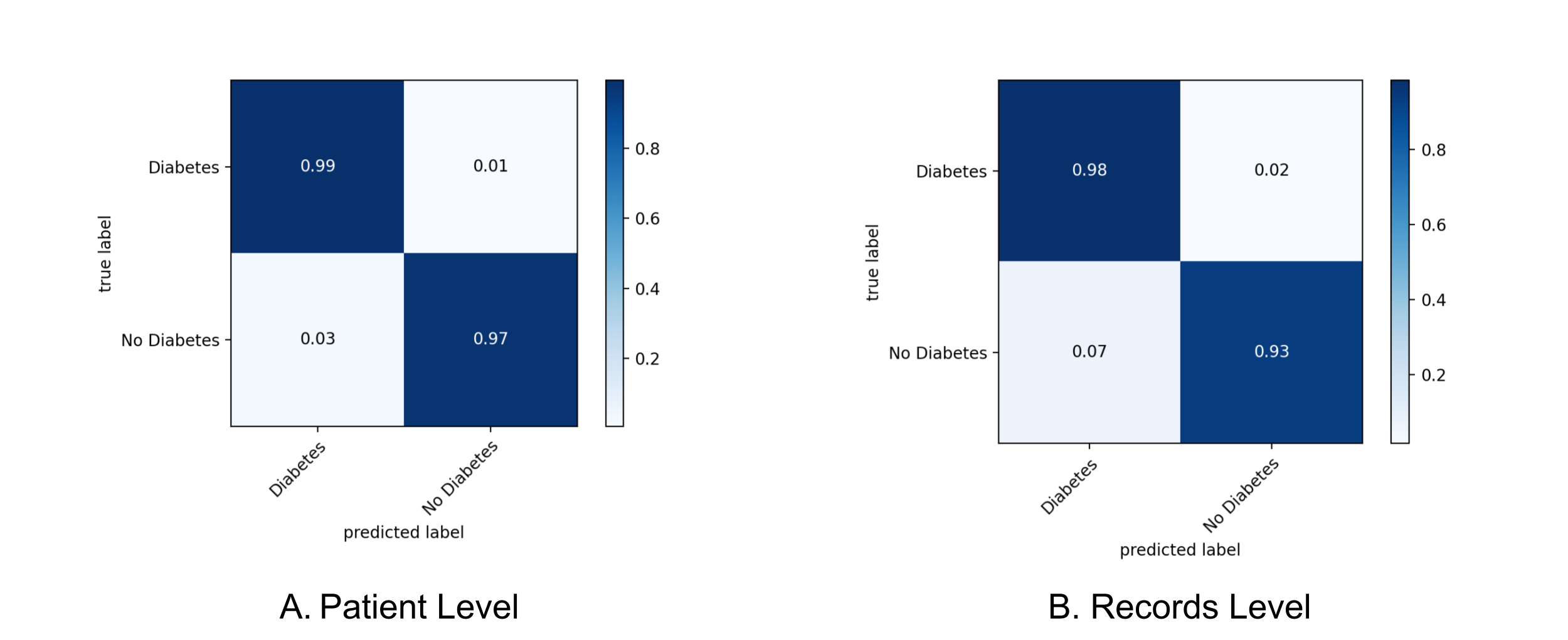}
   \caption{Confusion Matrix.}
   \label{fig:figure6}
\end{figure}

Figure~\ref{fig:figure3}A depicts the standard full-attention method, wherein each token is compared with every other token in a sequence. This method maximizes the information captured at the expense of computational efficiency. Figure~\ref{fig:figure3}B presents sparsity attention, another attention mechanism, which chunks the input sequences into blocks to reduce the token size and computational complexity. It represents an existing effort to apply the block patterns of fixed strides to sparsify the attention matrix. Figure~\ref{fig:figure3}C describes the time-based sparsity attention, in which the frequency of tokens is temporal and weights are more heavily assigned to tokens that are closer and less so to those farther away.

Figure~\ref{fig:figure3}D shows our TSA. It implements a fixed-patch aggregation on a new dimension Y to compose a 2D numerical representation of the PPG waveforms. The existing dimension X carries a series of numerical values representing time-sequence waveforms with a patch width of T. Since the second dimension, Y, is also time-based, we named this the Time Square Attention (TSA). Our tokenization method groups adjacent points into a square (2D) shape. The extended coverage includes sizes like 2 by 2, 3 by 3, 4 by 4, 5 by 5, etc. This approach effectively reduces the volume of tokens required to process by more than 10 times. For example, a 10 min-long PPG waveform at 128 Hz would comprise of about 77K sampled points. This optimization significantly increases computational efficiency.

TSA is essential in effectively tokenizing points so the Transformer can analyze longer sequences, while retaining both local representation and global contextual information. TSA uses 2D tokenization to link both short- and long-distance points, embedding connections within each token. It calculates the relationship of each token to every other token in the X and Y dimensions.

\subsection{Dynamic Patch Sizes in the 2D Transformer}

One of the key challenges in TSA is defining the optimal patch size used to form the 2D representation, maximizing the relationship analysis for all contexts. We explore a series of patch sizes to generate a group of dynamical 2D representations in different dimensions (Figure~\ref{fig:figure4}). Through dynamic patching, the various time dependencies are connected to address relationships between different distances, which are analyzed in parallel to learn the best performing patch size of TSA (forming the optimized shape of the patches).

Since each 2D patch can be processed as a 2D tensor representation, we simply apply the existing 2D Transformer algorithms to perform the “image classification” training. In our research, we deployed the hierarchical Swin, capturing both local and global dependencies within the 2D representations.

A detailed approach to the generating of different-sized 2D representations is explained in Algorithm~\ref{alg:algorithm1}. In our experiment, we marked T as 1024 points, representing 8s of the PPG waveforms.

\subsection{A Gated Network of Mixture-of-Experts}

To optimize the model performance from the dynamic patches in TSA, we then deploy the hierarchical structures of the patches in dynamic sizes and propose a gated mixture-of-experts (MoE) network, as demonstrated in Figure~\ref{fig:figure2}. 

An ensemble function is chosen as follows: 

\[ y = \sum_{i=1}^{N}{G(x)_i * E_i (x)} \]

\[ G(x) = Softmax(x * W_g)\]

where \({y}\) denotes the final diabetes prediction score, \({x}\) represents the PPG input, \({G}\) represents the output of the gating network, \({E}\) represents the output of the expert network, \({N}\) denotes the total number of the experts, and \({W_g}\) represents a trainable weight matrix.

A group of 2D representations of the different shapes are computed in each TSA, which connects with an MLP layer for the diabetes classifier and generates a likelihood estimation score via softmax. Within the MoE learning process, each expert's weight is computed. Then, the weights are combined to determine the final diabetes-detection outcome from these models.

Our proposed MoE approach successfully learns more specialized and disentangled expert features from the dynamic 2D patches. This enables HDformer to combine the complementary information from the different patch sizes while reducing the interference between them. Incorporating MoE, HDformer yields the best classification performance via its parallel, ensemble learning. For our study’s purposes, we took the configuration of five TSA modules with the dynamic patch sizes T (8 seconds), 2T, 4T, T/2, T/4.

\begin{figure}[t]
  \centering
  \includegraphics[width=0.9\linewidth]{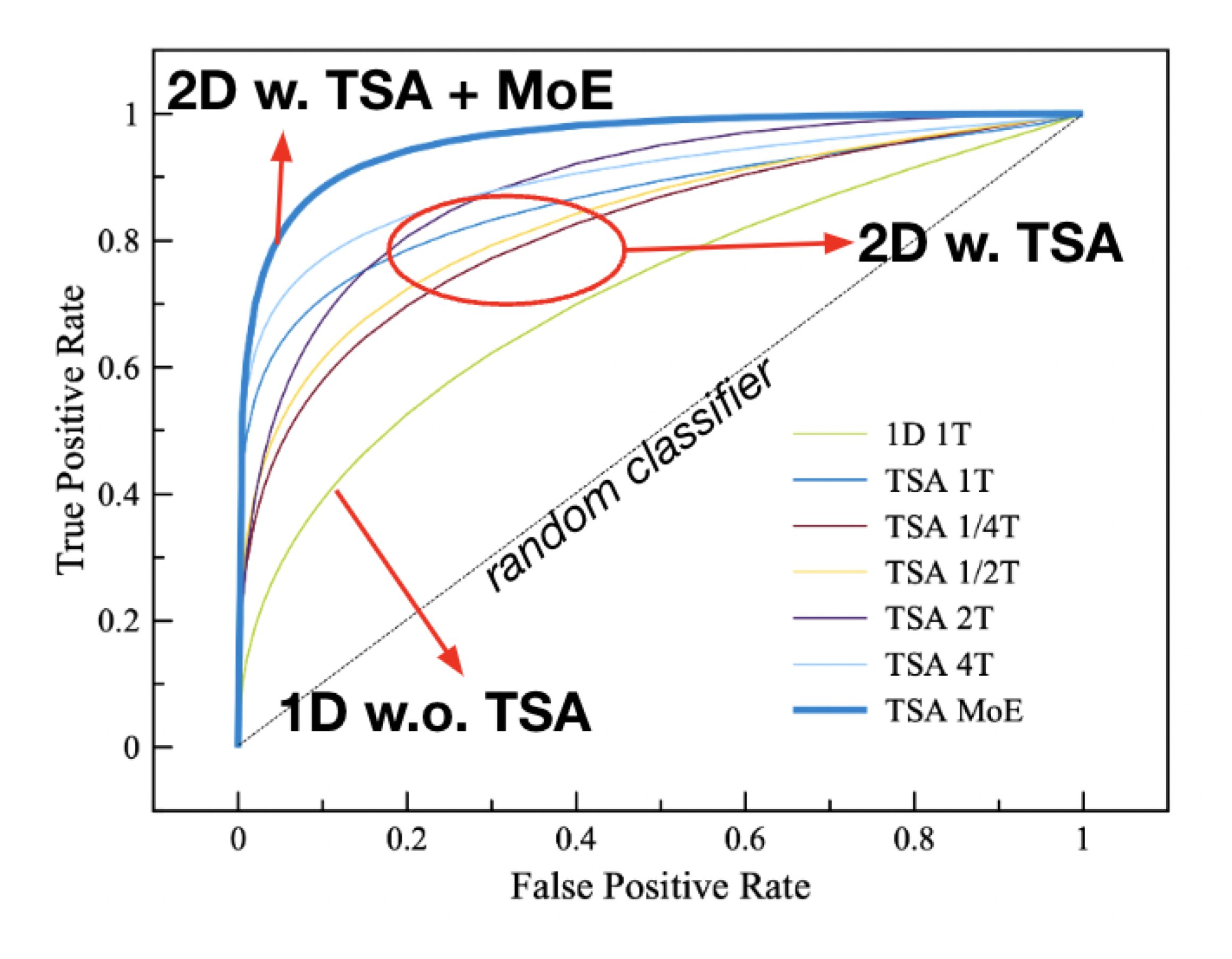}
   \caption{TSA Area Under the Curve Analysis.}
   \label{fig:figure7}
\end{figure}

\section{Experiments and Discussion}

\subsection{Datasets and Environment}

\begin{table}
  \begin{center}
  \begin{tabular}{@{} c c c c c @{}}
    \toprule
    Attentions & Sensitivity & Accuracy & Specificity & AUC \\
    \midrule
    1D 8s & 75.8 & 72.5 & 68.9 & 0.791\\
    1D 30s & 78.1 & 74.5 & 70.5 & 0.806\\
    1D 60s & 71.5 & 68.8 & 65.6 & 0.728\\
    TSA 30s & 77.8 & 75.1 & 72.5 & 0.808\\
    TSA 60s & 81.9 & 79.1 & 78.5 & 0.815\\
    TSA 180s & 83.2 & 81.5 & 80.9 & 0.829\\
    TSA 6m & 88.1 & 85.9 & 85.8 & 0.891\\
    TSA 10m & \textbf{98.4} & \textbf{97.3} & \textbf{92.8} & \textbf{0.929}\\
    \bottomrule
  \end{tabular}
  \end{center}
  \caption{Signals Lengths Study.}
  \label{tab:leng}
\end{table}


We used the public dataset MIMIC-III ~\cite{johnson2016mimic}, a comprehensive single-center database covering 38,597 distinct adult patients admitted to critical care units in a large tertiary care hospital. This dataset includes vital signs (like PPG and ECG), medications, laboratory measurements, procedure codes, diagnostic codes (ICD9 codes starting with 250 are labeled as diabetic patients), imaging reports, etc. One of the major reasons for choosing MIMIC-III was to evaluate our model in a standard comparison, rather than in a self-collected private dataset. All PPG waveforms were resampled to 128 Hz, and regular denoising and normalization were performed as part of the pre-processing. HDformer was implemented via PyTorch, and the model was trained on AWS instances with a NVIDIA A10G GPU.

\subsection{Evaluation}

We performed the evaluation by generating the confusion matrix on both the records level and the patient level, shown in Figure~\ref{fig:figure6}, in which the result from the patient level is an aggregated conclusion for all PPG results under the same individual. A higher accuracy at the patient level compared to the individual record level suggested real-world applications.

The model performed with an accuracy higher than 95\%, significantly outperforming previous research. As explained in Table~\ref{tab:sota}, HDformer achieved SOTA results on the MIMIC-III when evaluated on sensitivity, accuracy, specificity, and AUC. This finding reflects the benefits of long-range PPG processing with its effective TSA tokenization.

\begin{table}
  \begin{center}
  \begin{tabular}{@{} c c c c c @{}}
    \toprule
    Attentions & Sensitivity & Accuracy & Specificity & AUC \\
    \midrule
    1D 1T & 75.8 & 72.5 & 68.9 & 0.791\\
    TSA 1T & 86.8 & 85.6 & 82.9 & 0.879\\
    TSA T/4 & 78.9 & 77.5 & 75.2 & 0.835\\
    TSA T/2 & 81.8 & 79.9 & 78.0 & 0.858\\
    TSA 2T & 87.5 & 86.2 & 84.5 & 0.890\\
    TSA 4T & 89.6 & 89.1 & 87.8 & 0.895\\
    TSA MoE & \textbf{98.4} & \textbf{97.3} & \textbf{92.8} & \textbf{0.929}\\
    \bottomrule
  \end{tabular}
  \end{center}
  \caption{TSA Model Configuration Comparison.}
  \label{tab:comp}
\end{table}

We experimented with the existing long-range Transformers (discussed in the Related Work) to process the 1D PPG waveforms, and Informer ~\cite{zhou2021informer} yields the best results. Therefore,  Informer is taken as the default model to perform all the 1D analysis in the following discussion.

A comparison among the models with 1D, 2D with TSA, and 2D with TSA + MoE is plotted in the AU). HDformer (2D with TSA + MoE) achieved the best classification as depicted by the ROC and AUC, shown in Figure~\ref{fig:figure7}, demonstrating the different contributions from TSA and MoE to the final results.

The experiments suggest the effectiveness of HDformer and TSA through their novel designs. Our solution can efficiently analyze long-range PPG signals to accurately classify diabetes. Using the proposed TSA self-attention to aggregate a new dimension and the gated MoE layer to concatenate expert predictions, HDformer captures the key relationships between and within waveforms of dynamic patch sizes.

\subsection{TSA in Depth and Ablation Study}

To understand the effects of the different sizes of TSA and the MoE, we performed an ablation study on the different parameter configurations of the model.

\begin{table}
  \begin{center}
  \begin{tabular}{@{} c c c c c @{}}
    \toprule
    Attentions & Sensitivity & Accuracy & Specificity & AUC \\
    \midrule
    Image 8s & 61.8 & 59.9 & 57.8 & 0.678\\
    Image 30s & 65.9 & 62.8 & 59.5 & 0.686\\
    Image 60s & 58.1 & 56.5 & 53.7 & 0.645\\
    Image 180s & 52.9 & 51.9 & 50.8 & 0.618\\
    TSA 30s & 77.8 & 75.1 & 72.5 & 0.808\\
    TSA 60s & 81.9 & 79.1 & 78.5 & 0.815\\
    TSA 180s & 83.2 & 81.5 & 80.9 & 0.829\\
    TSA 6m & 88.1 & 85.9 & 85.8 & 0.891\\
    TSA 10m & \textbf{98.4} & \textbf{97.3} & \textbf{92.8} & \textbf{0.929}\\
    \bottomrule
  \end{tabular}
  \end{center}
  \caption{2D Tensor vs 2D Image Representation.}
  \label{tab:1d2dpv}
\end{table}

\subsubsection{The Impact of the Long-range and TSA}

\begin{table}
  \begin{center}
  \begin{tabular}{@{} c c c c c @{}}
    \toprule
    Attentions & Sensitivity & Accuracy & Specificity & AUC \\
    \midrule
    TSA ViT & 96.8 & 95.9 & 91.9 & 0.910\\
    TSA Swin & \textbf{98.4} & \textbf{97.3} & \textbf{92.8} & \textbf{0.929}\\
    \bottomrule
  \end{tabular}
  \end{center}
  \caption{Swin vs ViT on TSA.}
  \label{tab:vision}
\end{table}

To evaluate the effects of the long-range PPG on diabetes detection, a sensitivity analysis of different PPG waveform lengths is presented in Table~\ref{tab:leng}. For 1D sequences, performance metrics enhanced when the wavelength increased from 8s to 30s, as a result of adding more features and extending long-distance dependencies into the training. Interestingly, the continuous increase from 30s to 60s diluted the performance due to computational overload. Via TSA, the processing of 1D waveforms via 2D representations significantly reduced the size of the tokens without compromising long-term or short-term dependencies. The increase in wavelength then consistently improved the performance, illustrating the value of long-range PPG while using TSA to optimize the computation capacity.

\subsubsection{The Impact of the Dynamic Patching with MoE}

We experimented with different PPG inputs using different model parameter configurations, shown in Table~\ref{tab:comp}. The 2D representation in the TSA helped to achieve better results than the original 1D waveform in the standard self-attention mechanism. It is interesting to find that the larger size patches (2T and 4T) performed better than smaller patches (T/2 and T/4). The additional ensemble network from the gated MoE also yielded a considerable enhancement to the model's performance.

\begin{figure}[t]
  \centering
  \includegraphics[width=0.9\linewidth]{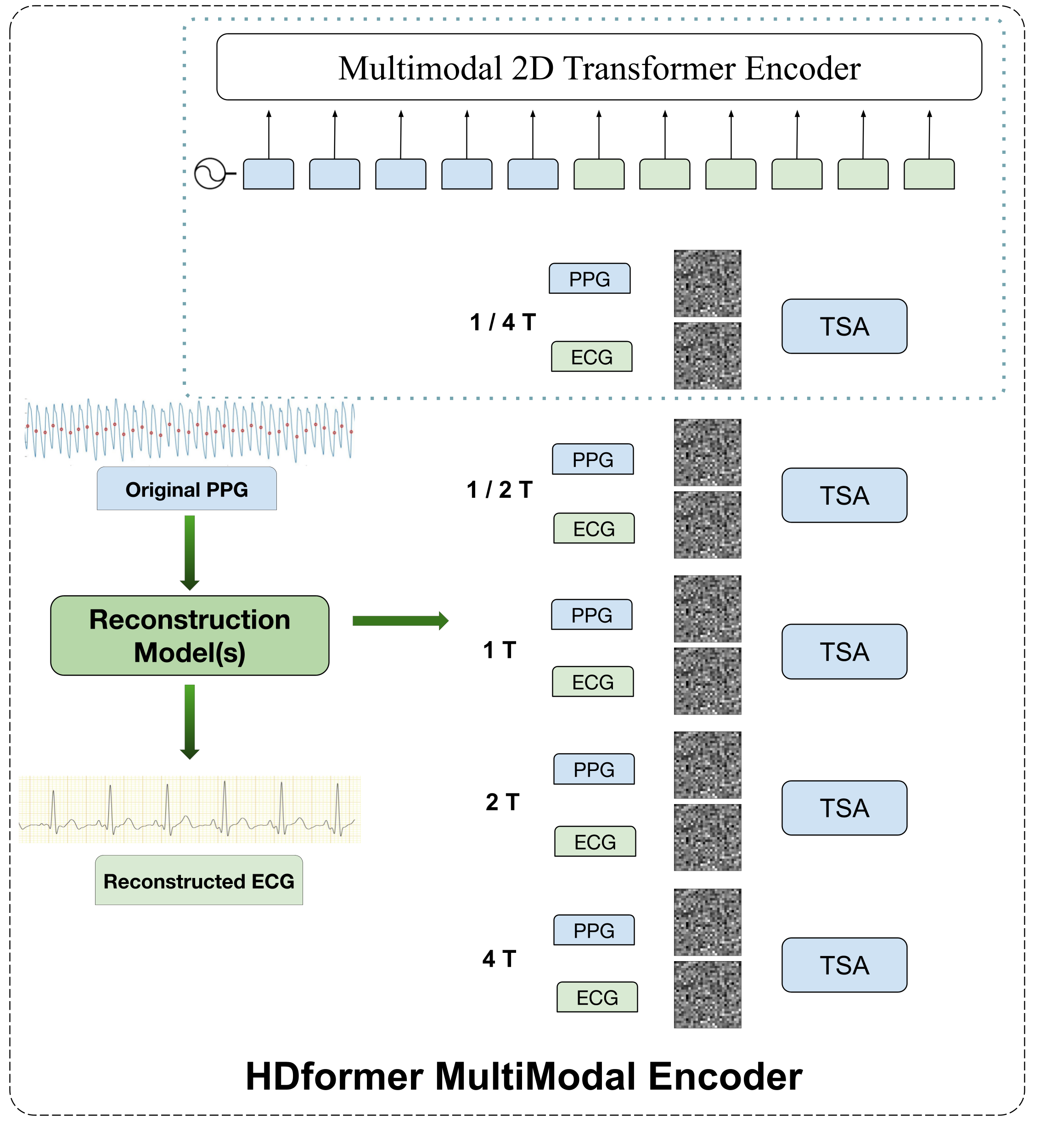}
   \caption{HDformer Extension.}
   \label{fig:figure9}
\end{figure}

\subsubsection{The Impact of the 2D Numerical Tensor}

TSA takes in numerical tensors; each time-based sample point is represented by a single value in 1D. For the 2D tokens, the respective values of adjacent points are concatenated into a single 2D tensor and then processed by a 2D Transformer. 



Numerical tensors enable greater efficiency than do image tensors. We compared the existing TSA to the 2D image-based representations which convert 1D PPG data into 2D images and found that the 2D image-based representations reduced the efficiency of long-range data processing by introducing more tokens (pixels) than did the 1D PPG (Table~\ref{tab:1d2dpv}).

\subsubsection{The Comparison of Existing 2D Transformers}

To evaluate the different Vision Transformer algorithms on the 2D TSA representations, we also compare the results between the standard ViT and the hierarchical Swin Transformer, as illustrated in Table~\ref{tab:vision}. While ViT performed with high accuracy, the Swin Transformer achieved better results. We hypothesize this is caused by the hierarchical structure of Swin, which captures the longer-distance dependencies of the 2D PPG with its different window sizes.

\begin{table}
  \begin{center}
  \begin{tabular}{@{} c c c c c @{}}
    \toprule
    Attentions & Sen. & Acc. & Spe. & AUC \\
    \midrule
    5m & 85.1 & 82.9 & 81.8 & 0.861\\
    5m + 256hz & 87.9 & 86.1 & 85.9 & 0.875\\
    5m + 256hz + MM & \textbf{95.9} & \textbf{95.1} & \textbf{91.5} & \textbf{0.917}\\
    10m & \emph{98.4} & \emph{97.3} & \emph{92.8} & \emph{0.929}\\
    \bottomrule
  \end{tabular}
  \end{center}
  \caption{TSA Exploration on Higher Frequency and Mulit-Modal.}
  \label{tab:products}
\end{table}

\subsection{HDformer Extension}

To validate the generalizability of the HDformer, we extend our experiments onto another commonly used dataset \cite{liang2018new} for diabetes detection and achieve the accuracy 98.9 with AUC 0.955. 

In addition to PPG, electrocardiography (ECG) is another broadly used digital biomedical waveform in clinics. ECGs provide important information about the electrical activity of the heart and are considered the gold standard for diagnosing many heart conditions as well as evaluating cardiovascular disease. Furthermore, ECGs can provide important clues about cardiac complications and cardiovascular disease burdens in patients with diabetes.

Recent research has yielded promising results in reconstructing the ECGs from PPGs \cite{zhu2019ecg} \cite{sarkar2021cardiogan} \cite{lan2023performer}. As such, we extend the HDformer architecture to incorporate the multi-modal inputs from both PPG and its reconstructed ECG, applying TSA to both waveforms (Figure~\ref{fig:figure9}).

\begin{figure}[t]
  \centering
  \includegraphics[width=0.9\linewidth]{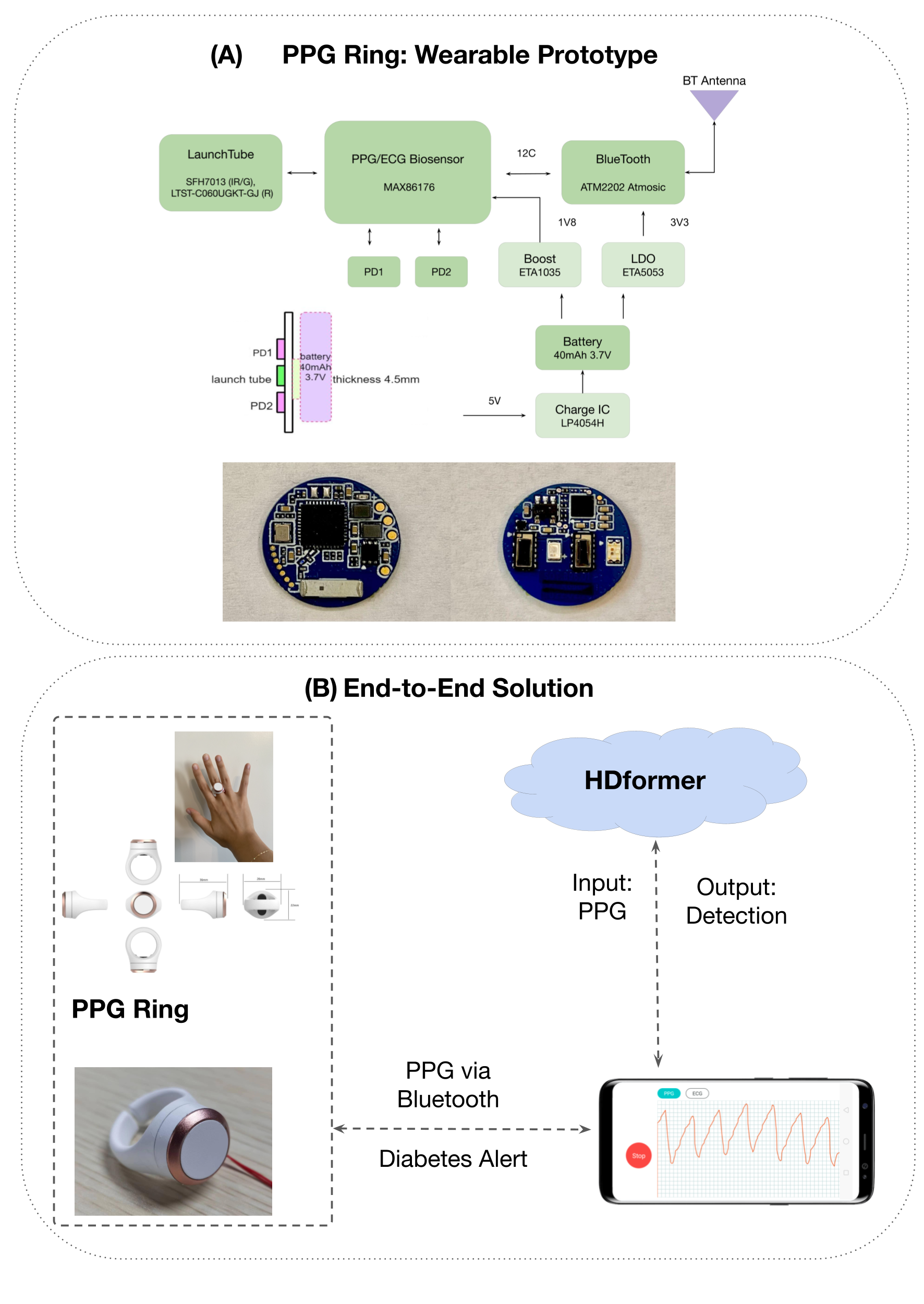}
   \caption{An End-to-End Solution with PPG Rings.}
   \label{fig:figure10}
\end{figure}

The multi-modal TSA shortened the required long-range PPG wavelength to 5 min, which yielded a comparable performance to the 10 min PPG waveform (Table~\ref{tab:products}).

The sampling frequency defines the volume of PPG data per second. Higher frequencies yield more data to be processed. We compared the different frequences and found that although 256 Hz increased the model performance marginally for the same length of the PPG, simply increasing the frequency of the PPG did not make up for the accuracy deficit caused by the decrease in waveform length.

Combined with the optimization from the higher frequency, the multi-modal HDformer extension achieved high performance with only 5 min of PPG. Although still considered long-range and still requiring for optimized computational efficiency, this methods could provide a more practical solution in the real-world.

Beside the diabetes detection, the HDformer extension also demonstrates a high performance (95+ accuracy) on the CVD detection during our experiment on the CVD labelled PPG waveforms from the MIMIC-III dataset, including coronary artery disease (CAD), congestive heart failure (CHF), myocardial infarction (MI), and hypotension (HOTN). 

\subsection{Medical Applications}

With greater efficiency in enabling long-range PPG processing, HDformer represents a way of monitoring and detecting diabetes in a non-invasive, scalable way. Given that PPG, the only raw input into the HDformer, is low-cost and user-friendly to retrieve, we prototyped a PPG-based ring wearable, as presented in Figure~\ref{fig:figure10}. We developed this ring as a proof of concept for our model and applied it in a real-world setting.

We host the trained HDformer model in the Cloud. There, HDformer takes in the PPG waveforms from the wearable and infers the 2D representations as part of the process of predicting diabetes. The wearable rings are convenient to wear, allowing for the consistent collection of long-range PPG signals and easy adoption for most users.

Our end-to-end solution enables a scalable, convenient, and affordable approach to detecting diabetes via PPG. This solution can provide instantaneous detection for individuals, help doctors easily screen for diabetes, and safeguard underprivileged communities, ultimately helping the 240 million undiagnosed people to receive treatment and preserve their lives \cite{ogurtsova2022idf}.

\begin{figure}[t]
  \centering
  \includegraphics[width=0.9\linewidth]{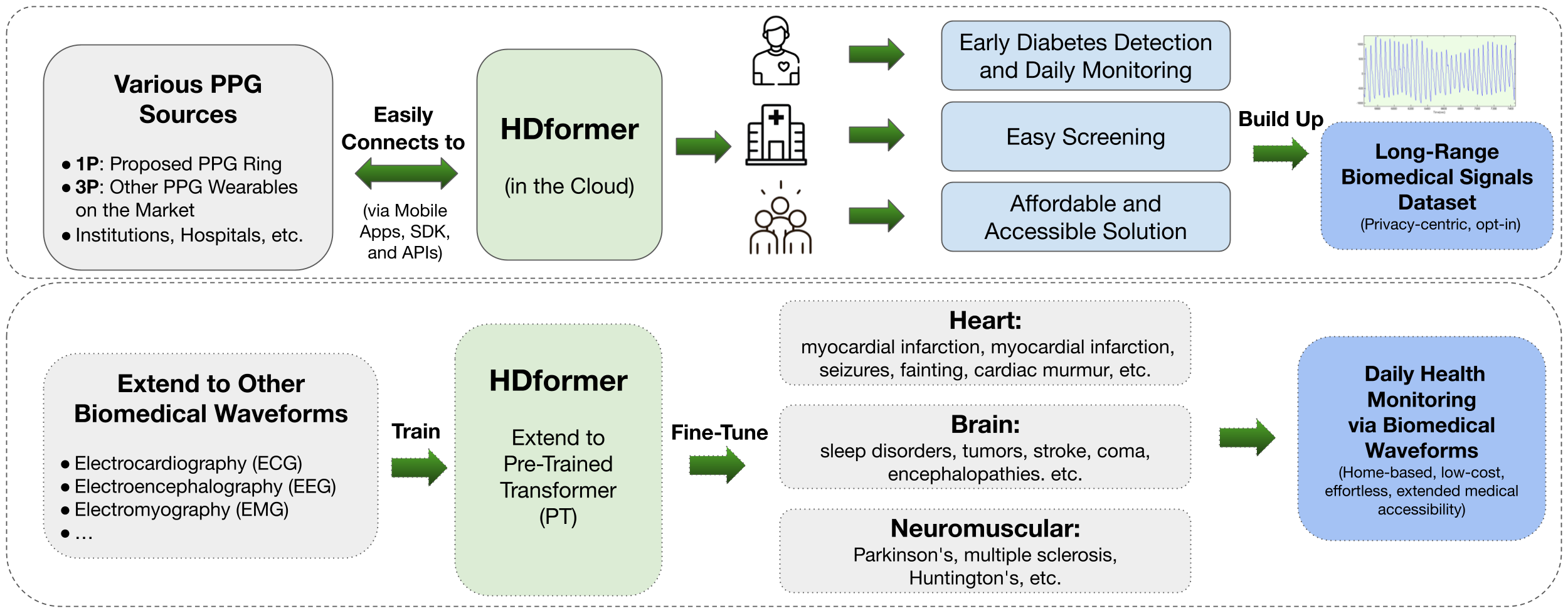}
   \caption{A HDformer Ecosystem.}
   \label{fig:figure11}
\end{figure}

In addition, HDformer can serve as an ecosystem to support various PPG wearables and institutions which carry PPG data, creating a new, privacy-centric, opt-in-based long-range PPG dataset, building up a large biomedical pre-trained model, to benefit research communities, as depicted in Figure~\ref{fig:figure11}.

\section{Conclusion}

We propose HDformer, a Transformer-based model capable of processing long-range vascular signals of PPG to predict diabetes. Our model achieves SOTA performance, enabling a novel, non-invasive approach to early diabetes detection, and is suitable for widespread clinical application. The proposed TSA module demonstrates high efficiency in processing long-range data with 2D representations and a gated MoE layer helps ensemble the classification from the dynamic patch sizes of the 2D TSA. This method is adaptable to other long-range biomedical waveforms, paving the way for the use of such signals in diverse disease screening and management. Furthermore, the method establishes a foundational, pre-trained Transformer that can serve as the basis for future large-scale model refinements, especially for the general time-series sequences processing.


{\small
\bibliographystyle{ieee_fullname}
\bibliography{egbib}
}

\end{document}